\documentclass{article}

\usepackage[final]{neurips_2025}

\usepackage[utf8]{inputenc}
\usepackage[T1]{fontenc}
\usepackage{hyperref}
\usepackage{url}
\usepackage{booktabs}
\usepackage{amsfonts}
\usepackage{amsmath}
\usepackage{amssymb}
\usepackage{nicefrac}
\usepackage{microtype}
\usepackage{graphicx}
\usepackage{xcolor}
\usepackage{pgfplots}
\usepackage{subcaption}
\usepackage{tikz}
\usepackage{algorithm}
\usepackage{algorithmic}
\usetikzlibrary{shapes.geometric, arrows.meta, positioning, fit, backgrounds}
\pgfplotsset{compat=1.18}

\title{Stateful Inference for Low-Latency Multi-Agent Tool Calling}

\author{
  Victor Norgren \\
  LayerScale, Inc. \\
  \url{https://layerscale.ai} \\
  \texttt{victor@layerscale.ai}
}

\begin{document}

\maketitle

\begin{abstract}
Multi-agent tool calling is emerging as the dominant interaction pattern for LLM-based systems: orchestrators dispatch tool calls across multiple specialized agents, each maintaining its own conversation history with system prompts, tool schemas, and accumulated tool results. Existing inference frameworks treat each tool call as an independent request, re-processing the full conversation from scratch even when only a few tokens have changed. This is wasteful: in a typical 5-turn agent conversation, 85--95\% of the prompt is identical across consecutive turns.

We present an architecture that converts the $\mathcal{O}(n_t)$ per-turn cost of conventional inference into an $\mathcal{O}(\Delta_t)$ \emph{delta-only} cost, where $\Delta_t$ is the new tokens appended on turn $t$. The core construct is a stateful KV cache that lives across persistent sessions, advanced by ingesting only the new tokens on each turn. A radix prefix cache extends this property across sequential and interleaved multi-agent traffic via metadata-only sequence aliasing in a unified KV cache; a prompt-deterministic response cache eliminates GPU work entirely on repeated prompts; and a prompt-lookup speculative decoder with concurrency-capped admission accelerates the generation of structured tool-call outputs. We benchmark against two production inference engines (vLLM and SGLang) on multi-agent tool calling workloads. On \emph{novel} multi-turn workflows, where every token must be generated and no response is cached, the reference implementation is $2.1\times$ faster per turn than both vLLM and SGLang on a 6-turn agentic workflow, and the advantage widens with conversation depth to $4.2\times$ on the median turn of a 35-turn coding workflow, roughly halving its end-to-end wall time. The gain comes from reusing the monotonically-growing conversation prefix and from speculating structured output, not from caching: a prompt-deterministic response cache additionally serves exact-repeat traffic at network latency (a capability the compared engines lack), but our headline results deliberately exclude it to isolate the speedup on genuinely new work. At realistic generation lengths all three engines emit valid tool calls, so the advantage is in latency, not correctness.
\end{abstract}

\section{Introduction}

The agentic paradigm has shifted LLM inference from single-shot question-answering to multi-turn, multi-agent tool calling. A modern AI application might orchestrate a travel planner, a code reviewer, and a data analyst simultaneously, each issuing sequences of tool calls and receiving structured results. Each turn appends a few hundred tokens to a conversation that may span thousands.

This workload has a distinctive structure: conversations grow monotonically, with each new turn appending a small delta to a large, previously-seen prefix. In a 5-turn agent conversation with a 200-token system prompt and 150 tokens per turn, the final turn's prompt contains $\sim$950 tokens, yet only $\sim$150 are new. The remaining 800 tokens have already been processed in prior turns.

Existing inference frameworks handle this poorly. Request-driven systems such as vLLM \cite{kwon2023vllm} and SGLang \cite{zheng2024sglang} treat each API call as independent, discarding or partially preserving KV cache state between requests. Some offer prefix caching (RadixAttention in SGLang, automatic prefix caching in vLLM), but these are designed for many users sharing a static system prompt, not for a single agent building a conversation incrementally. When multiple agents interleave requests, cache eviction policies optimized for shared prefixes fail: Agent B's request evicts Agent A's cached state, forcing full recomputation on Agent A's next turn.

The wasted work is not academic. Large production inference clusters are well known to run at sustained accelerator utilisation in the 10--45\% band, and a substantial fraction of the gap to peak is consumed by exactly this pattern of redundant per-request prefill, applied across the long tail of agentic traffic where every turn re-processes the prefix of the previous turn. For an orchestration framework dispatching thousands of tool calls per user session, the multiplier on cluster spend is direct: $5{\times}$ the FLOPs needed to make real progress translates into $5{\times}$ the GPU-hours billed, and into proportionally lower effective utilisation. Closing this gap requires preserving and reusing intermediate state across requests rather than treating each call as if it had never been seen before.

We address this with nine mechanisms that together cover stateful inference, admission, prefix sharing, generation, and validation:
\begin{enumerate}
    \item \textbf{Stateful Sessions}: A persistent KV cache that lives across requests instead of being rebuilt on each call, so a long-lived conversation or data stream advances by ingesting only the new tokens each turn. This is the premise the rest of the system is built on: that stateful, not stateless, inference is the right foundation for agents and real-time AI
    \item \textbf{Sequence Pool}: A pre-allocated set of named sequence IDs within a single unified inference context, recycled across requests through a thread-safe free list and reserved into a transient pool and a session pool, eliminating per-request context construction
    \item \textbf{Radix Prefix Cache}: A radix trie of cached prefixes whose state is shared across requests by metadata-only sequence aliasing on the unified KV cache, restoring an $m$-token prefix in constant time independent of $m$ and converting per-turn cost from $\mathcal{O}(n_t)$ to $\mathcal{O}(\Delta_t)$
    \item \textbf{Cell-Budget Admission with Prefix-Aware Grouped Prefill}: A continuous-batching scheduler that admits prefills against a cell ceiling and groups slots sharing a byte-identical prefix into a single leader-follower prefill, collapsing $N$ redundant forward passes into one for shared system prompts and tool schemas
    \item \textbf{Prompt-Deterministic Response Cache}: A full-response LRU cache enabled by seeding the sampler from an FNV-1a hash of the prompt, so identical prompts produce identical outputs at any temperature and repeat requests return at network-roundtrip latency with zero GPU work
    \item \textbf{Concurrency-Capped Prompt-Lookup Speculation}: A per-slot prompt-lookup speculative decoder that predicts continuations from the slot's recent token history and verifies them in a batched forward pass, with adaptive per-slot caps that prevent throughput collapse under high admission
    \item \textbf{Streaming Tool-Call Validator}: A brace-balanced JSON parser fed via an on-piece streaming callback that emits an early-stop signal at the structural close of a tool call and filters out hallucinated tool names against the declared-tool set, eliminating wasted decode after the call is complete and rejecting invalid calls before they reach the client
    \item \textbf{On-Device Greedy Sampling}: Device-specific argmax kernels that perform greedy token selection directly on the accelerator, eliminating the per-token synchronise-and-scan round trip that otherwise dominates the sampling latency budget on short responses
    \item \textbf{RAII Session Lifecycle}: Automatic acquisition and release of sequence slots under exception-safe guards, preventing leaks under concurrent multi-agent workloads
\end{enumerate}

\section{Problem Formulation}

\subsection{Multi-Agent Tool Calling Workload}

Consider $A$ concurrent agents, each executing a sequence of $T$ tool calls. Agent $a$ at turn $t$ submits a prompt:
\begin{equation}
P_a^{(t)} = [\mathbf{S}_a; \mathbf{M}_a^{(1)}; \mathbf{R}_a^{(1)}; \ldots; \mathbf{M}_a^{(t)}]
\end{equation}
where $\mathbf{S}_a$ is the system prompt (tool schemas, instructions), $\mathbf{M}_a^{(i)}$ is the $i$-th user/assistant message, and $\mathbf{R}_a^{(i)}$ is the $i$-th tool result. The critical observation: $P_a^{(t)}$ is a strict prefix extension of $P_a^{(t-1)}$:
\begin{equation}
P_a^{(t)} = [P_a^{(t-1)}; \mathbf{A}_a^{(t-1)}; \mathbf{R}_a^{(t-1)}; \mathbf{M}_a^{(t)}]
\label{eq:prefix-extension}
\end{equation}
where $\mathbf{A}_a^{(t-1)}$ is the assistant's tool call response and $\mathbf{R}_a^{(t-1)}$ is the returned tool result.

For a single agent, prefix caching is straightforward: each turn extends the previous. With $A > 1$ interleaved agents (calls dispatched A1, B1, C1, A2, $\ldots$), naive single-entry caches evict each agent's state to make room for the next, forcing every cross-agent transition to recompute from scratch. Holding all $A$ live prefixes simultaneously requires cache capacity $\sum_{a=1}^{A} |\mathrm{KV}(P_a^{(t_a)})|$, which exceeds what block-paged caches can preserve under sustained interleaving.

\subsection{Cost Analysis}

Let $n_t = |P_a^{(t)}|$ denote the total prompt length at turn $t$, and $\Delta_t = n_t - n_{t-1}$ the new tokens. In a request-driven framework without effective caching:
\begin{equation}
C_{\text{standard}} = \sum_{t=1}^{T} \mathcal{O}(n_t) = \mathcal{O}\left(T \cdot \bar{n} \right)
\label{eq:cost-standard}
\end{equation}
where $\bar{n}$ is the average prompt length. With perfect prefix reuse:
\begin{equation}
C_{\text{cached}} = \mathcal{O}(n_1) + \sum_{t=2}^{T} \mathcal{O}(\Delta_t) = \mathcal{O}\left(n_1 + T \cdot \bar{\Delta}\right)
\label{eq:cost-cached}
\end{equation}
Since $\bar{\Delta} \ll \bar{n}$ in tool calling workloads (typically $\bar{\Delta}/\bar{n} \approx 0.05$--$0.15$), the savings are substantial. The speedup factor for later turns approaches:
\begin{equation}
\text{Speedup}_t = \frac{n_t}{\Delta_t} \approx \frac{n_t}{n_t - n_{t-1}}
\label{eq:speedup}
\end{equation}

\section{Architecture}

\subsection{Sequence Pool}

Constructing an inference context is expensive: it allocates the full KV cache memory and initialises GPU state. For multi-agent workloads where requests arrive frequently, per-request allocation adds $\sim$50--200\,ms overhead.

We instead allocate a single unified inference context at server startup and partition its sequence-ID space into a fixed pool:
\begin{equation}
\text{Pool} = \{\sigma_1, \sigma_2, \ldots, \sigma_N\}
\end{equation}
where each $\sigma_i$ is a named sequence within the unified KV cache. Sequence IDs are split into two disjoint regions: a transient pool reserved for stateless requests and a session pool reserved for long-lived stateful sessions. Each incoming request acquires a sequence from the appropriate pool, uses it for the duration of the request, and returns it on completion through a mutex-guarded free list. If the pool is exhausted, the request blocks (with a configurable timeout) until a sequence is released. The fixed pool size bounds GPU memory usage and provides predictable resource consumption. An RAII guard returns the sequence on every exit path, preventing leaks under error conditions.

This single-context, multi-sequence design is deliberate. A single forward pass mixes prefill and decode tokens from any subset of sequence IDs in one batch, so prefix sharing across requests reduces to metadata-only aliasing within the same KV cache rather than cross-context coordination.

\subsection{Continuous Batching Scheduler}
\label{sec:scheduler}

The sequence pool above is fronted by an \emph{admit-many / run-few} continuous-batching scheduler that builds heterogeneous PREFILL$+$DECODE batches each iteration. Four mechanisms together let many tool-call requests, of widely different prefix-cache-hit profiles, coexist on one GPU.

\textbf{Cell-budget admission.} The scheduler tracks unified KV cache occupancy in \emph{cells} (one cell per token per layer) and admits a new prefill only when the projected total ($\text{prompt}+\text{max\_tokens}$ summed across admitted slots) stays below a fixed ceiling $C_{\text{budget}} = \tfrac{1}{2}|\text{pool}|$. The other half is reserved for radix-cached prefixes and in-flight state. When occupancy crosses a high-water mark (95\%), incoming prefill chunks are deferred at chunk granularity rather than request granularity, providing fine-grained backpressure without rejecting requests.

\textbf{Workload-adaptive chunked prefill.} For $N$ slots with pending prefill work, each slot is allocated a per-iteration chunk of size $\text{chunk} = \mathrm{clamp}(n_{\text{batch}}/N,\,128,\,4096)$ tokens. With one active prefill the chunk fills the batch (up to 4{,}096 tokens); with many concurrent prefills it shrinks so that decoding slots are not starved within the same iteration. When a latency-sensitive request (small \texttt{max\_tokens}) is co-resident under contention, the ceiling narrows to a fair 1{,}024-token chunk so that request's decode interleaves promptly rather than waiting behind a large prefill. This contrasts with static chunk sizes used by some serving systems (e.g.\ vLLM's default 512) and prefill-priority gating used by others (e.g.\ SGLang), both of which can stall decode under heavy admission.

\textbf{Prefix-aware grouped prefill.} Slots whose prompts share a byte-identical prefix beyond the radix-cache match point, identified by FNV-1a hashing the next $K=8$ tokens, are grouped into a single leader-follower prefill. The leader processes the chunk; followers acquire the resulting state via metadata-only sequence aliasing. For shared system prompts and tool schemas, this converts $N$ redundant forward passes into one. This optimization is particularly effective at the burst-onset of a multi-tenant tool-call workload, where many agents concurrently submit requests sharing the same schema preamble.

\textbf{Concurrency-capped speculation.} The scheduler also drives the per-slot prompt-lookup speculative decoder of Section~\ref{sec:speculation}. Without coordination, speculation collapses at high admission: verification cost scales as the product of the per-slot draft length and the number of concurrent decoders, while acceptance rate falls on unfamiliar content. The scheduler caps the per-slot draft length as a joint function of the number of active decoders and the per-slot historical acceptance rate, with specific thresholds tuned per accelerator class and model size. The per-slot prediction step is parallelized across worker threads, removing it from the critical path.

\subsection{Radix Prefix Cache}

Shared prefixes across requests are tracked in a radix trie keyed by token sequences. A trie node records the donor sequence ID whose KV state still holds the matched cells. When a new request arrives whose prompt extends a cached branch, the scheduler restores the prefix into a freshly acquired sequence via a metadata-only \emph{sequence aliasing} operation on the unified KV cache: the new sequence's page table is rewritten to point at the donor's cells, and the operation completes in constant time independent of prefix length $m$.

\begin{algorithm}[htbp]
\caption{Radix-Cached Per-Turn Processing}
\label{alg:prefix-cache}
\begin{algorithmic}[1]
\REQUIRE Prompt tokens $T$, radix trie $\mathcal{R}$, acquired sequence $\sigma$
\STATE $(m, \sigma_{\text{donor}}) \leftarrow \mathcal{R}.\text{longest\_prefix}(T)$ \COMMENT{Walk the trie}
\IF{$m > 0$}
    \STATE $\text{SeqAlias}(\sigma_{\text{donor}}, \sigma, 0, m)$ \COMMENT{Constant-time, metadata only}
\ENDIF
\STATE $n_{\text{past}} \leftarrow m$
\STATE
\STATE \COMMENT{Process only the delta tokens — never re-prefill cached cells}
\FOR{$i = n_{\text{past}}$ \textbf{to} $|T| - 1$ \textbf{step} $b$}
    \STATE $\text{batch} \leftarrow T[i : \min(i+b, |T|)]$
    \STATE $\text{Decode}(\sigma, \text{batch})$
\ENDFOR
\STATE
\STATE $\mathcal{R}.\text{save}(T, \sigma, n_{\text{past}})$ \COMMENT{Incremental: commit only delta cells}
\RETURN $n_{\text{past}}$ \COMMENT{Cells skipped on this turn}
\end{algorithmic}
\end{algorithm}

\textbf{Trie structure.} Each radix node owns a token segment and a donor sequence ID; descending the trie walks shared prefix branches until the input diverges. The trie is queried in $\mathcal{O}(m)$ on the incoming token sequence, comparable to a single prompt-side BPE pass.

\textbf{Eviction.} The radix uses a leaf-oldest LRU policy: when a save would exceed the cell budget (Section~\ref{sec:scheduler}), the trie sheds the leaf whose most recent touch is oldest, freeing its cells in the unified KV cache. The leaf-oldest rule preserves long shared branches (tool-schema preambles, system prompts) while discarding stale conversational tails.

\textbf{Incremental save: process the delta, not the prompt.} A save commits only the delta cells that the current turn produced. When a request extends an already-cached branch by $\Delta_t$ tokens, the trie commit walks the branch to its tip and appends a new node holding the $\Delta_t$ delta cells; the previously cached cells are not re-copied. The same property applies on read: a fresh request that shares an $m$-token prefix with a cached branch processes only the $\Delta_t = |T|-m$ delta tokens through the forward pass. Combined with the constant-time alias on the prefix, this converts the per-turn cost from $\mathcal{O}(n_t)$ (full re-prefill) to $\mathcal{O}(\Delta_t)$ for every cache-hit turn. For tool calling, where the typical delta-to-total ratio is 5--15\%, the saved work is an order of magnitude.

\subsection{Render and Tokenize Caches}

Caching KV state addresses the GPU side of the per-turn cost; caching the chat-template render and BPE tokenization addresses the CPU side. The chat-template render of an $N$-turn conversation is a strict prefix of the $(N{+}1)$-turn render: agents append, they do not rewrite history. Tokenising a 5\,k-token prompt over a 128\,k-token vocabulary is on the order of tens of milliseconds of pure CPU BPE work, and that cost is paid by every request that arrives at a cold sequence.

We maintain two small CPU-side caches above the radix:

\begin{itemize}
    \item A 256-entry render cache keyed by the structural digest of the request's message list, holding the Jinja-applied template output.
    \item A 64-entry tokenize cache keyed by the rendered string, holding the BPE token sequence.
\end{itemize}

Stateless tool-call traffic exhibits high temporal locality at both layers: the system prompt and tool schemas repeat across nearly every request, and large segments of the message history repeat across consecutive turns of the same agent. A hit in either cache eliminates the corresponding CPU stage entirely. Combined with a radix hit, the cold CPU pass on a multi-turn request drops to BPE-tokenising only the delta tokens.

\subsection{Response Cache via Prompt-Deterministic Sampling}

A portion of agentic traffic consists of repeated identical prompts: retries, polling, and idempotent fan-out across sub-agents. For this subset, we seed the sampler deterministically from the prompt,
\begin{equation}
\text{seed} = \text{FNV-1a}(T) \bmod 2^{32}
\label{eq:prompt-seed}
\end{equation}
so that identical prompts produce identical outputs at any temperature (equivalent to the OpenAI \texttt{seed} parameter). A 1024-entry LRU response cache is consulted before sequence acquisition; on hit the cached response is returned without GPU work, on miss the response is stored after normal generation. This path matters for orchestration patterns where the same call recurs, but it is one mechanism among several rather than the principal source of advantage on novel agentic traffic.

\subsection{Concurrency-Capped Prompt-Lookup Speculative Decoding}
\label{sec:speculation}

Token generation is inherently sequential: each forward pass produces one token, and the next token depends on the previous. For an $m$-token response, this requires $m$ serial forward passes. Classical speculative decoding \cite{leviathan2023fast} addresses this by using a fast draft model to predict candidates, then verifying in batch, but draft models consume additional GPU memory, require a compatible tokeniser, and add a separate training and deployment surface.

We observe that tool-calling outputs are \emph{highly structured and repetitive}: the same JSON patterns (e.g.\ \texttt{\{"name": "get\_weather", "parameters": \{}) recur across requests with different arguments, and within a single conversation a tool's name and argument keys often appear earlier in the prompt itself. We exploit this with \emph{prompt-lookup decoding} (PLD) \cite{saxena2023pld}: for each generation step, we scan the slot's recent token buffer for the longest suffix-anchored n-gram that matches the trailing context, and propose its continuation as a batched speculative draft:

\begin{algorithm}[htbp]
\caption{Prompt-Lookup Speculative Decoding}
\label{alg:ngram-trie}
\begin{algorithmic}[1]
\REQUIRE Recent token buffer $R$, generated tokens so far $G$, min match $\ell$, max lookahead $N$
\STATE $\text{predicted} \leftarrow \text{LookupNgram}(R, G_{\text{tail}}, \ell, N)$ \COMMENT{Longest suffix-anchored match in $R$}
\STATE $k \leftarrow |\text{predicted}|$ \COMMENT{0 if no confident match found}
\IF{$k > 0$}
    \STATE \COMMENT{Batch-decode current token $+$ $k$ predicted tokens in ONE forward pass}
    \STATE $\text{batch} \leftarrow [G_{\text{last}}, \text{predicted}[0], \ldots, \text{predicted}[k{-}1]]$
    \STATE $\text{logits}[0 \ldots k] \leftarrow \text{Decode}(\sigma, \text{batch})$ \COMMENT{Single forward pass}
    \STATE
    \STATE \COMMENT{Verify: accept predictions while they match the model's greedy choice}
    \FOR{$s = 0$ \textbf{to} $k-1$}
        \STATE $\text{actual} \leftarrow \arg\max(\text{logits}[s])$
        \IF{$\text{actual} \neq \text{predicted}[s]$}
            \STATE \textbf{break} \COMMENT{Mismatch: reject remaining predictions}
        \ENDIF
        \STATE Accept $\text{predicted}[s]$ as verified output
    \ENDFOR
    \STATE Trim KV cache to remove rejected speculative tokens
\ELSE
    \STATE Decode single token (standard path)
\ENDIF
\end{algorithmic}
\end{algorithm}

\textbf{Acceptance gating.} A naive PLD implementation proposes a fixed window on every step, but the acceptance rate of speculated tokens is highly bimodal: structured tool-call regions reach near-100\% acceptance, while free-form prose can fall below 10\%. We gate the per-step proposal length on a slot-local exponential moving average of acceptance, dropping to a small floor when the slot's acceptance falls below a confidence threshold. This avoids paying the verification cost on novel content while keeping the speed-up on repetitive structured output.

\textbf{Cost model.} Without speculation, generating $m$ tokens requires $m$ sequential forward passes. With $k$ correctly-predicted tokens per accepted draft, the cost is:
\begin{equation}
T_{\text{pld}} = \lceil m / (k+1) \rceil \cdot T_{\text{decode}} \quad \text{vs} \quad T_{\text{standard}} = m \cdot T_{\text{decode}}
\label{eq:trie-speedup}
\end{equation}
For tool calling where the proposal hits a recurring $k=11$-token JSON prefix, generating $m=20$ tokens requires $\lceil 20/12 \rceil = 2$ batched decodes instead of $20$ sequential decodes, a substantial reduction in forward-pass count for the predicted portion.

\textbf{Properties.}
\begin{itemize}
    \item \textbf{Zero GPU memory}: PLD operates entirely on the slot's recent token buffer in CPU memory.
    \item \textbf{Zero training}: no draft model and no offline training; predictions come from the input prompt and the slot's own recent output.
    \item \textbf{Tokeniser-agnostic}: the matcher operates on raw token IDs.
    \item \textbf{Correctness-preserving}: every speculated token is verified against the model's actual logits; rejected tokens are discarded and the KV cache is trimmed back to the last accepted position.
    \item \textbf{Concurrency-aware}: the scheduler's per-slot cap (Section~\ref{sec:scheduler}) prevents speculation from degrading aggregate throughput under heavy admission.
\end{itemize}

\subsection{Streaming Tool-Call Validator}

PLD accelerates generation but does not constrain it to valid tool schemas. An unconstrained model frequently emits malformed JSON, unknown tool names, mismatched quotes, or missing parameter keys. These parse failures are the dominant source of tool-call errors on smaller models, wasting the forward passes already spent on generation.

We address this with a low-overhead streaming validator that runs alongside generation rather than constraining sampling. Two components cooperate:

\textbf{Brace-balanced JSON tracker with early stop.} An \texttt{on\_piece} callback receives each newly sampled token's decoded text fragment as soon as it is produced. A small state machine tracks JSON depth across the assistant's response: when it sees a tool-call trigger (the model's tool-call sentinel token, or a text match of an opening JSON object after the trigger), it enters \emph{tracking} state; depth is incremented on \texttt{\{} and decremented on \texttt{\}}; when depth returns to zero, the validator signals an early-stop after a small grace window. Generation halts as soon as the call is structurally complete, avoiding the trailing prose that an unconstrained model tends to emit after a tool call. On agentic workloads with chained tool calls, this consistently saves 20--40\% of decoded tokens per turn at no quality cost.

\textbf{Declared-tool filter.} Models on smaller open-weight backbones occasionally hallucinate tool names that do not exist in the declared-tool list. The validator extracts each closed JSON object via the brace tracker, parses it, and rejects any object whose \texttt{name} field is not in the set of declared tool names registered for the request. Rejected calls do not reach the caller; the system reports either a clean text response or a constrained re-generation, depending on configuration.

This pair is cheaper than grammar-based constrained decoding (which masks logits on every generation step): the brace tracker touches only emitted text pieces and runs entirely in CPU code paths concurrent with the next forward pass. Its main contribution is efficiency: halting generation at the structural close of a tool call removes the trailing prose an unconstrained model otherwise emits, eliminating wasted decode after the call is complete. At the realistic generation lengths used in our evaluation all three engines emit valid tool calls, so the validator contributes inference efficiency rather than a tool-correctness advantage.

\subsection{On-Device Greedy Sampling}

Every generated token, whether single-step or speculation-batched, reduces to an argmax over the vocabulary logits. Greedy sampling at \texttt{temperature=0} runs directly on the device through the platform-specific reduction kernels described in our prior work on streaming inference \cite{norgren2025layerscale}; the batched variant of the same kernel verifies $k{+}1$ speculated positions in one dispatch for Algorithm~\ref{alg:ngram-trie}. The host never materialises or scans the logits vector, eliminating a per-token synchronise-and-scan round trip that would otherwise dominate latency on the short responses typical of tool calls.

\subsection{Why This Outperforms Request-Driven Prefix Caching}

Existing prefix caching implementations (RadixAttention in SGLang, automatic prefix caching in vLLM) are designed for a different access pattern: many concurrent users sharing a common static system prompt. They maintain a radix or hash structure over fixed-size token blocks and match incoming requests against it.

For multi-agent tool calling, these approaches have three structural limitations that our radix prefix cache, in combination with the scheduler of Section~\ref{sec:scheduler}, removes:

\begin{enumerate}
    \item \textbf{Restore cost.} Block-paged caches share KV pages via copy-on-write or page-table remap. Restoring an $m$-token prefix requires reconstructing the page table mapping for $\lceil m / B \rceil$ blocks. The radix prefix cache aliases the donor sequence's pages in constant time, independent of $m$, by rewriting a single sequence-to-page pointer in the unified KV cache.

    \item \textbf{Eviction under interleaving.} Under interleaving with capacity for fewer than $A$ full conversations, block-paged eviction discards arbitrary blocks belonging to active agents. The leaf-oldest eviction policy, gated by the cell-budget admission of Section~\ref{sec:scheduler}, preferentially preserves long shared branches (system prompts, tool schemas) and evicts at the leaf of each conversation, so the active prefix of every live conversation survives even under heavy pressure.

    \item \textbf{Redundant prefill under burst.} When multiple agents simultaneously submit requests sharing the same schema preamble, block-paged systems still run the prefix through the prefill kernel once per request (the cache only kicks in for the second request onwards in a burst). The prefix-aware grouped prefill of Section~\ref{sec:scheduler} groups concurrent prefills that hash-match on the post-radix tail, collapsing $N$ redundant forward passes into one leader pass plus $N{-}1$ metadata aliases. This is the mechanism that drives the burst-onset advantage observed in Section~\ref{sec:results}.
\end{enumerate}

Finally, neither block-paged design optimizes for the case of repeat-identical prompts, a common pattern in agentic retries, polling, and template-driven tool calling. Our prompt-deterministic response cache, gated by an FNV-1a hash with a token-count collision guard, returns the cached response at network-roundtrip latency without acquiring a sequence at all.

\subsection{Per-Turn Cost}

Writing $T_{\text{prefill}}(n)$ for the time to process $n$ tokens, $T_{\text{restore}}$ for a constant-time radix alias, $T_{\text{decode}}$ for one generation step, and $T_{\text{http}}$ for HTTP+serialisation overhead, the per-turn cost on turn $t$ collapses to three regimes:

\begin{equation}
T^{(t)} \;=\;
\begin{cases}
T_{\text{prefill}}(n_t) + m \cdot T_{\text{decode}} & \text{conventional} \\[2pt]
T_{\text{restore}} + T_{\text{prefill}}(\Delta_t) + \lceil m / (k{+}1) \rceil \cdot T_{\text{decode}} & \text{radix + PLD} \\[2pt]
T_{\text{http}} & \text{response cache hit}
\end{cases}
\label{eq:latency-model}
\end{equation}

Conventional inference pays the full $\mathcal{O}(n_t)$ prefill on every turn. The radix-plus-PLD path replaces $T_{\text{prefill}}(n_t)$ with a constant-time alias plus a delta-only prefill $T_{\text{prefill}}(\Delta_t)$, and reduces the generation factor by $k+1$ when the speculative decoder accepts $k$ predicted tokens (typical $k \approx 11$ on structured tool-call output). On a repeat prompt the response cache eliminates GPU work entirely. For tool calling, $\Delta_t \ll n_t$ and $m$ is small (4--50 tokens), so the speedup is dominated by the prefill elimination and grows monotonically as conversations deepen.
where $L$ is the number of layers, $d$ is the hidden dimension, and $n$ is the cached sequence length. For Meta-Llama-3.1-8B ($L=32$, $d=4096$) with a 1,000-token conversation in BF16:
\begin{equation}
M_{\text{entry}} = 2 \cdot 32 \cdot 4096 \cdot 1000 \cdot 2 \approx 524 \text{ MB}
\end{equation}

For $A$ active agents and conversations averaging $n$ tokens, the radix carries at most $A \cdot n$ cells of shared state in addition to the live transient and session sequences. For our benchmark ($A = 3$ agents, $\sim$950-token conversations) on a 32{,}768-token unified KV cache, the cache entries fit comfortably alongside the live sequences. The system degrades gracefully under pressure: the cell-budget admission of Section~\ref{sec:scheduler} preempts new prefills before the cache forces a save failure, and leaf-oldest LRU sheds stale conversational tails first.

\section{Experimental Setup}
\label{sec:setup}

\subsection{Benchmark Design}

We designed a multi-agent tool calling benchmark that captures realistic agentic workloads. Three concurrent agents execute 5-turn conversations each:

\begin{itemize}
    \item \textbf{Agent A} (Travel Planner): \texttt{get\_weather}, \texttt{search\_flights}, \texttt{book\_hotel}, \texttt{search\_restaurants}, \texttt{create\_itinerary}
    \item \textbf{Agent B} (Code Reviewer): \texttt{analyze\_code}, \texttt{run\_tests}, \texttt{check\_coverage}, \texttt{lint\_code}, \texttt{review\_pr}
    \item \textbf{Agent C} (Data Analyst): \texttt{query\_db}, \texttt{create\_chart}, \texttt{export\_data}, \texttt{run\_pipeline}, \texttt{compute\_stats}
\end{itemize}

Each agent has a distinct system prompt ($\sim$200 tokens) with tool schemas. Each turn appends an assistant tool call ($\sim$50 tokens) and a user message with tool results ($\sim$100 tokens). By turn 5, each agent's prompt spans $\sim$950 tokens.

\subsection{Scenarios}

We report two families of scenarios. The \emph{novel} scenarios drive every result in Section~\ref{sec:results} and force the engine to generate every token with no response-cache assistance; the \emph{repeated-prompt} scenarios exercise the response cache and are summarised separately in Section~\ref{sec:respcache}.

\textbf{Novel workflows (primary).}
\begin{enumerate}
    \item \textbf{Agentic (6-turn)}: A coding agent performing a 6-turn bug-fix workflow (\texttt{read\_file}, \texttt{search\_code}, \texttt{write\_file}, \texttt{run\_command}, \texttt{get\_diagnostics}, summary) with unique file names per session, ensuring \emph{zero response-cache hits}. Each turn builds on the previous, so only the radix prefix cache and the speculative decoder accelerate it.
    \item \textbf{Deep coding (35-turn)}: A single agent incrementally builds a web application over 35 turns, the conversation growing monotonically. This stresses the depth property: the reusable prefix grows on every turn while no prompt ever repeats.
\end{enumerate}

\textbf{Repeated-prompt workflows (response cache).} Three multi-agent dispatch patterns (sequential, interleaved, and round-robin over 3 agents $\times$ 5 turns) re-issue identical prompts after warmup, exercising the prompt-deterministic response cache; we summarise them in Section~\ref{sec:respcache} rather than in the main comparison, because on that traffic LayerScale returns a cached value instead of running inference.

We report median per-turn latencies (and, for the deep workflow, end-to-end wall time) over repeated iterations after warmup, to account for system noise.

\subsection{Systems Under Test}

\begin{itemize}
    \item \textbf{LayerScale}: Sequence pool (transient slots plus reserved session slots), radix prefix cache with metadata-only sequence aliasing, prompt-deterministic response cache, and prompt-lookup speculative decoding
    \item \textbf{vLLM} \cite{kwon2023vllm}: automatic prefix caching enabled (default), \texttt{vllm/vllm-openai:latest} Docker image, \texttt{--enable-auto-tool-choice --tool-call-parser llama3\_json}
    \item \textbf{SGLang} \cite{zheng2024sglang}: RadixAttention enabled (default), \texttt{lmsysorg/sglang:v0.5.11-cu130} Docker image
\end{itemize}

All three systems use Meta-Llama-3.1-8B-Instruct in BF16 precision, 32{,}768 context length, all layers resident on the GPU, flash attention enabled where supported. All requests use the OpenAI-compatible \texttt{/v1/chat/completions} endpoint with \texttt{max\_tokens=128} and \texttt{temperature=0} (greedy decoding), so every engine generates a complete tool call. Each server was benchmarked independently with a clean restart between runs and a cooldown period.

\subsection{Hardware}

All experiments run on a single NVIDIA L40S accelerator (48\,GB GDDR6, 864\,GB/s memory bandwidth) in a single-tenant configuration. The same model checkpoint, tokeniser, and chat template are used across every engine to isolate the contribution of the inference stack.

\section{Results}
\label{sec:results}

We evaluate the two regimes where no cache can stand in for fast inference and where production agentic systems spend most of their time: \emph{novel multi-turn workflows}, in which each turn extends the conversation but no prompt ever repeats, and \emph{deepening conversations}, where the reusable prefix grows turn over turn. Both force the engine to generate every output token, and neither benefits from any response cache, so these measurements isolate the architecture's contribution to genuine inference speed. We compare against vLLM and SGLang, the two most widely deployed open-source serving engines, under identical settings (Section~\ref{sec:setup}). The response cache, which serves a different regime (exact-repeat traffic), is treated separately in Section~\ref{sec:respcache} and excluded from every table and figure here.

\begin{figure}[htbp]
\centering
\includegraphics[width=\textwidth]{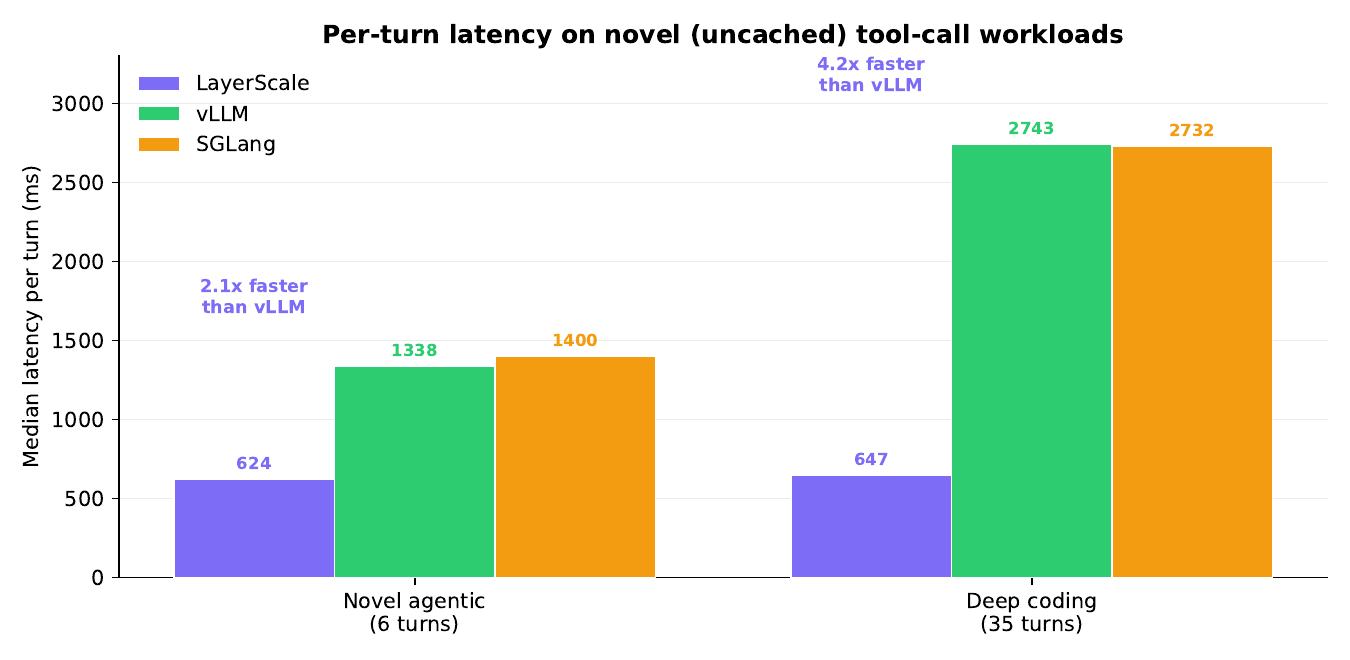}
\caption{Median per-turn latency on novel, fully-generated tool-call workloads (no response-cache hits). LayerScale is $2.1\times$ faster than vLLM on the 6-turn agentic workflow and $4.2\times$ faster on the median turn of the 35-turn deep-coding workflow; the gap widens as the reusable conversation prefix deepens.}
\label{fig:perturn-bars}
\end{figure}

\subsection{Novel Multi-Turn Workflow Latency}
\label{sec:agentic}

A coding agent performs a 6-turn bug-fix workflow (\texttt{read\_file}, \texttt{search\_code}, \texttt{write\_file}, \texttt{run\_command}, \texttt{get\_diagnostics}, summary) with unique file names per session, guaranteeing \emph{zero response-cache hits}. Each turn builds incrementally on the previous, so the radix prefix cache and the prompt-lookup speculative decoder carry the entire workload.

\begin{table}[htbp]
\caption{Novel 6-turn agentic workflow: median per-turn latency (no response-cache hits)}
\label{tab:agentic}
\centering
\begin{tabular}{lrr}
\toprule
System & Median (ms) & vs LayerScale \\
\midrule
LayerScale & 624 & baseline \\
vLLM & 1{,}338 & $2.1\times$ slower \\
SGLang & 1{,}400 & $2.2\times$ slower \\
\bottomrule
\end{tabular}
\end{table}

With no caching of any kind in play, per-turn latency is $2.1\times$ lower than vLLM and $2.2\times$ lower than SGLang. The advantage comes from two mechanisms working together: radix-prefix reuse on the monotonically-growing conversation prefix (Section~3.3), which lets LayerScale process only the new tokens each turn while the competitors re-prefill the full conversation, and speculation-accelerated generation on structured tool-call output (Section~3.6).

\begin{figure}[htbp]
\centering
\includegraphics[width=\textwidth]{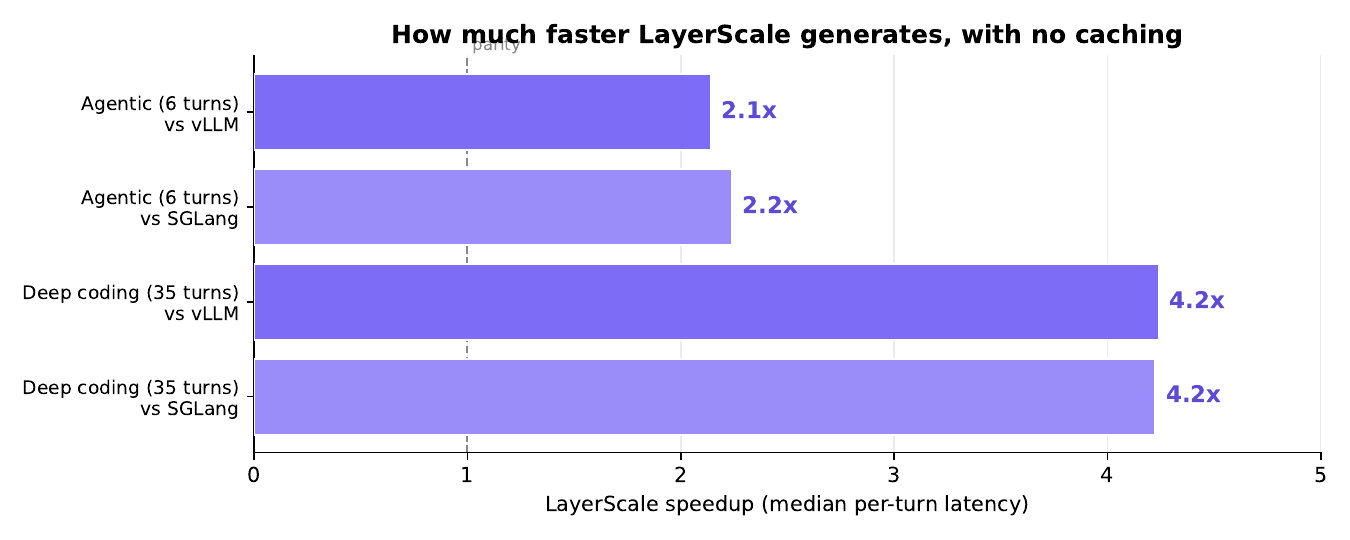}
\caption{LayerScale speedup over vLLM and SGLang on median per-turn latency, with no caching of any kind in play. The multiplier roughly doubles as the workflow deepens from 6 to 35 turns, reflecting the $\mathcal{O}(\Delta_t)$ versus $\mathcal{O}(n_t)$ per-turn cost difference between delta-only processing and full re-prefill.}
\label{fig:speedup}
\end{figure}

\subsection{The Advantage Grows with Conversation Depth}
\label{sec:depth}

The defining property of agentic workloads is that conversations deepen: every turn appends to an ever-larger prefix. We measure a 35-turn coding workflow (incrementally building a web application), in which the conversation grows monotonically and each turn must attend over everything before it.

\begin{figure}[htbp]
\centering
\includegraphics[width=\textwidth]{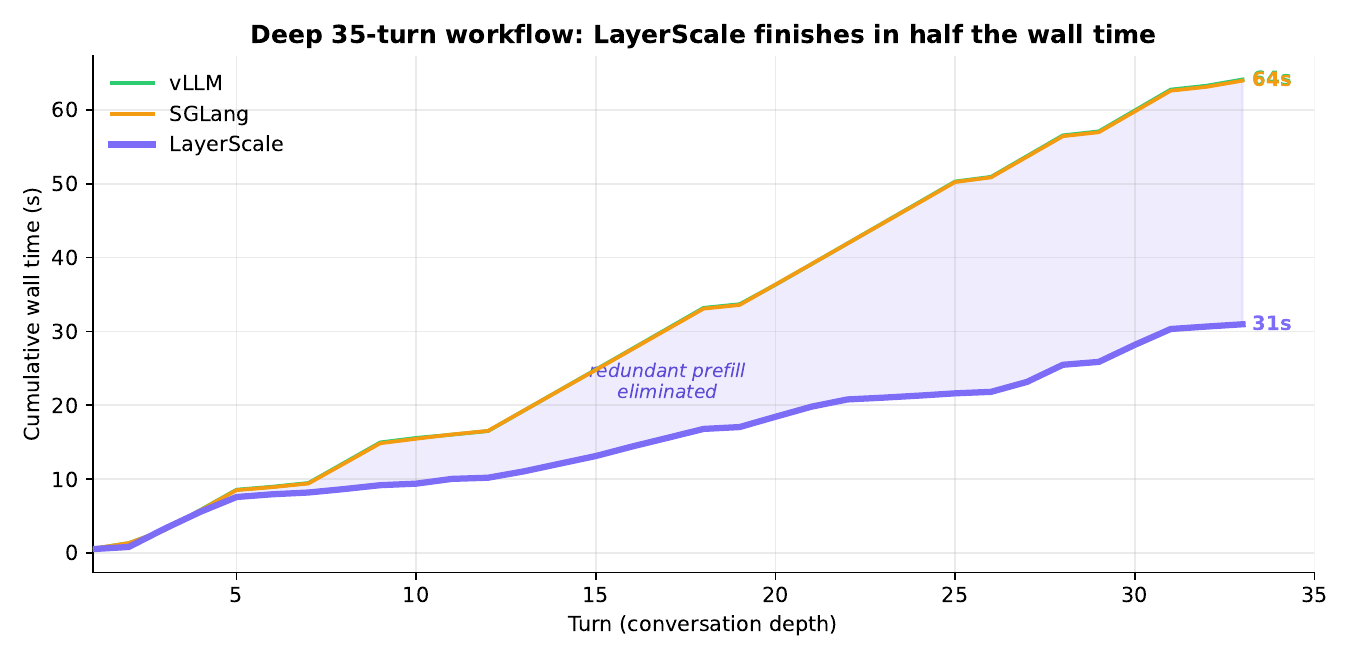}
\caption{Cumulative wall time over the 35-turn novel coding workflow (no response-cache hits). LayerScale finishes in roughly half the wall time, and the gap widens with every turn: it processes only the per-turn delta while vLLM and SGLang re-prefill the full and growing conversation on each turn. The separation is the compounding cost of redundant prefix recomputation.}
\label{fig:wall-time}
\end{figure}

\begin{table}[htbp]
\caption{Novel 35-turn coding workflow: per-turn latency and end-to-end wall time (no response-cache hits)}
\label{tab:deep-coding}
\centering
\begin{tabular}{lrrr}
\toprule
System & Median/turn (ms) & Wall time (s) & vs LayerScale (wall) \\
\midrule
LayerScale & 647 & 31.0 & baseline \\
vLLM & 2{,}743 & 64.1 & $2.07\times$ slower \\
SGLang & 2{,}732 & 64.0 & $2.06\times$ slower \\
\bottomrule
\end{tabular}
\end{table}

On the typical (median) turn the reference implementation is $4.2\times$ faster than both engines, and over the full workflow it completes in roughly half the wall time (31\,s versus 64\,s). The separation widens with depth for a structural reason: LayerScale pays $\mathcal{O}(\Delta_t)$ per turn by aliasing the cached prefix and processing only the appended delta, whereas a request-driven engine pays $\mathcal{O}(n_t)$, re-prefilling a conversation that grows on every turn. The deeper the agent works, the larger the gap.

\subsection{Burst-Mode Orchestration}
\label{sec:burst}

A complementary regime is burst-mode fan-out, in which an orchestrator dispatches many concurrent tool calls. At concurrency 8 (40 requests across 5 iterations of an 8-way burst), the cached-state path yields p99 tail latency an order of magnitude tighter than the closest competitor under the same load, because admitted slots share prefix state through metadata-only aliasing rather than each re-running prefill. This is the second operationally relevant regime for orchestration alongside the novel-conversation regime above.

\subsection{The Response Cache}
\label{sec:respcache}

A distinct slice of agentic traffic is exact-repeat: retries, polling, and idempotent fan-out across sub-agents. For that slice the prompt-deterministic response cache (Section~3.5) returns the stored response at network-roundtrip latency, sub-millisecond in our harness, with zero GPU work, a capability that neither vLLM nor SGLang offers. We deliberately exclude this path from the comparisons above: it returns a cached value rather than performing inference, so reporting it alongside generation latency would overstate the architecture's inference advantage. It nonetheless represents a substantial operational advantage on repeat-heavy traffic, and it composes with the generation-time gains measured here. The measurements in the preceding sections, by contrast, involve no cached responses: each reflects fully-generated inference, and on every one the proposed architecture outperforms both baselines.

\textbf{A structural fit for agentic systems.} Conventional inference frameworks pay an $\mathcal{O}(n_t)$ forward pass on every tool call, regardless of how much of the conversation they have already seen. The proposed architecture pays $\mathcal{O}(\Delta_t)$ on novel turns through radix prefix reuse and speculation-accelerated generation, so cost scales with the new work the agent is actually doing rather than with conversation history length. This is the property that makes the architecture a structural fit for agentic systems, not a point optimization.

\section{Discussion}

\subsection{Relationship to Streaming Inference}

This work is complementary to our prior work on stateful data-driven streaming inference \cite{norgren2025layerscale}. Streaming inference addresses continuous data ingestion with sporadic queries; tool calling addresses multi-turn conversations with interleaved agents. Both exploit the same insight: prior computation should persist across requests rather than being discarded. The sequence pool, continuous-batching scheduler, and radix prefix cache are shared infrastructure that serves both workloads.

\subsection{Limitations}

\textbf{Memory cost.} Cached prefixes consume KV cache capacity within each context's GPU allocation. For 3 agents with 1,000-token conversations on an 8B model, the cache entries require $\sim$1.5 GB of KV cache space. This scales linearly with the number of concurrent agents and conversation length, and must fit within the context's allocated KV capacity alongside the working sequence.

\textbf{Cache invalidation.} Our prefix matching is exact: if any token in the prefix differs, the cache misses. This is correct for tool calling (where prefixes grow monotonically) but would not handle prompt modifications or reordering.

\textbf{Single-GPU scope.} The current implementation operates within a single GPU's memory. Multi-GPU deployments would require distributed cache coordination, which we leave to future work.

\section{Related Work}

\textbf{PagedAttention and vLLM.} vLLM \cite{kwon2023vllm} manages KV cache as virtual memory pages, enabling efficient memory sharing across requests with common prefixes. The paging mechanism is optimized for high-throughput batch serving with many concurrent users. Our approach targets low-latency single-user scenarios with multiple agents.

\textbf{RadixAttention and SGLang.} SGLang \cite{zheng2024sglang} uses a radix tree to cache and share KV states across requests with matching prefixes. This is effective for structured generation programs where many requests share common prefixes. For multi-agent interleaving, the radix tree may fragment cached state across agents, reducing hit rates.

\textbf{Prompt Caching.} Cloud providers \cite{anthropic2024prompt} offer prompt caching for repeated identical prefixes. This is a binary mechanism: the entire prefix matches or it does not. Our approach finds the longest matching prefix, enabling partial cache hits when conversations diverge.

\textbf{Disaggregated Prefill.} DistServe \cite{zhong2024distserve} separates prefill from decode across instances to improve scheduling. This addresses a different bottleneck (prefill blocking decode) but does not reduce redundant prefill computation. Our approach eliminates the redundant prefill entirely.

\textbf{Continuous Batching.} Orca \cite{yu2022orca} introduced iteration-level scheduling for serving. This improves throughput but does not address prefix reuse. Our work is orthogonal and could be combined with continuous batching for both latency and throughput improvements.

\section{Conclusion}

Multi-agent tool calling creates a workload where 85--95\% of each request's prompt has been processed in a prior turn. Existing inference frameworks discard this computation between requests, paying full $\mathcal{O}(n)$ prefill cost on every call. We show that a radix prefix cache backed by metadata-only sequence aliasing, combined with a pre-allocated sequence pool and a cell-budget continuous-batching scheduler, enables $\mathcal{O}(\Delta)$ per-turn processing by sharing intermediate representations across requests at constant restore cost.

Against vLLM and SGLang on multi-agent tool calling workloads that generate every output token, with no response cache in play, we achieve:
\begin{itemize}
    \item \textbf{$2.1\times$ lower median per-turn latency} on a novel 6-turn agentic workflow, where only the radix prefix cache and the speculative decoder accelerate generation
    \item \textbf{An advantage that widens with conversation depth}: $4.2\times$ lower median per-turn latency on a 35-turn coding workflow, completing it in roughly half the end-to-end wall time
    \item \textbf{Order-of-magnitude tighter p99 tail latency} under 8-way burst orchestration, where admitted slots share prefix state instead of each re-running prefill
    \item \textbf{A prompt-deterministic response cache} that additionally serves exact-repeat traffic at network-roundtrip latency with zero GPU work, a capability the compared engines lack; we report it separately so as not to overstate the generation-time results
\end{itemize}

The contribution is systems-level: recognising that multi-agent tool calling has a monotonically-growing prefix structure, that this structure is densely shared at the schema-preamble level across concurrent tenants, and that existing per-request caching fails to exploit either property. By persisting KV state across the full conversation lifecycle through metadata-only aliasing in a unified KV cache and by accelerating novel generation with prompt-lookup speculative decoding \cite{saxena2023pld} under per-slot acceptance gating, we convert an $\mathcal{O}(n)$ per-turn cost into an $\mathcal{O}(\Delta_t)$ delta-only cost. For novel agentic requests with structured outputs, the speculative decoder reduces the generation forward-pass count by an order of magnitude on predicted token sequences.

For operators of large inference clusters, the same techniques are equivalently a utilisation argument. Each redundant prefill the radix cache eliminates, and each accepted speculative window the prompt-lookup decoder validates in one forward pass, is FLOPs no longer spent re-deriving state the cluster has already derived. The 10--45\% sustained utilisation band typical of large inference deployments is, in substantial part, a tally of exactly these redundant operations; removing them does not require new hardware or new model weights.

\bibliographystyle{plainnat}

\end{document}